\documentclass[10pt,twocolumn,letterpaper]{article}

\usepackage{cvpr}              

\usepackage{graphicx}
\usepackage{amsmath}
\usepackage{amssymb}
\usepackage{booktabs}

\usepackage[pagebackref,breaklinks,colorlinks]{hyperref}

\usepackage[capitalize]{cleveref}
\crefname{section}{Sec.}{Secs.}
\Crefname{section}{Section}{Sections}
\Crefname{table}{Table}{Tables}
\crefname{table}{Tab.}{Tabs.}

\def\cvprPaperID{7976} 
\def\confName{CVPR}
\def\confYear{2023}

\begin{document}

\title{
PAniC-3D: Stylized Single-view 3D Reconstruction \\
from Portraits of Anime Characters
}

\author{
    Shuhong Chen\textsuperscript{*\dag}
    \and Kevin Zhang\textsuperscript{*}
    \and Yichun Shi\textsuperscript{\dag}
    \and Heng Wang\textsuperscript{\dag}
    \and Yiheng Zhu\textsuperscript{\dag}
    \and Guoxian Song\textsuperscript{\dag}
    \and Sizhe An\textsuperscript{\dag}
    \and Janus Kristjansson\textsuperscript{*}
    \and Xiao Yang\textsuperscript{\dag}
    \and Matthias Zwicker\textsuperscript{*}
    \and \\
    University of Maryland - College Park, MD, USA\textsuperscript{*} \\
    ByteDance\textsuperscript{\dag} 
}
\maketitle


\begin{abstract}
   We propose PAniC-3D, a system to reconstruct stylized 3D character heads directly from illustrated (p)ortraits of (ani)me (c)haracters.  Our anime-style domain poses unique challenges to single-view reconstruction; compared to natural images of human heads, character portrait illustrations have hair and accessories with more complex and diverse geometry, and are shaded with non-photorealistic contour lines.  In addition, there is a lack of both 3D model and portrait illustration data suitable to train and evaluate this ambiguous stylized reconstruction task.  Facing these challenges, our proposed PAniC-3D architecture crosses the illustration-to-3D domain gap with a line-filling model, and represents sophisticated geometries with a volumetric radiance field.  We train our system with two large new datasets (11.2k Vroid 3D models, 1k Vtuber portrait illustrations), and evaluate on a novel AnimeRecon benchmark of illustration-to-3D pairs.  PAniC-3D significantly outperforms baseline methods, and provides data to establish the task of stylized reconstruction from portrait illustrations.
\end{abstract}


\begin{figure*}[t]
    \centering
    \includegraphics[width=1\linewidth]{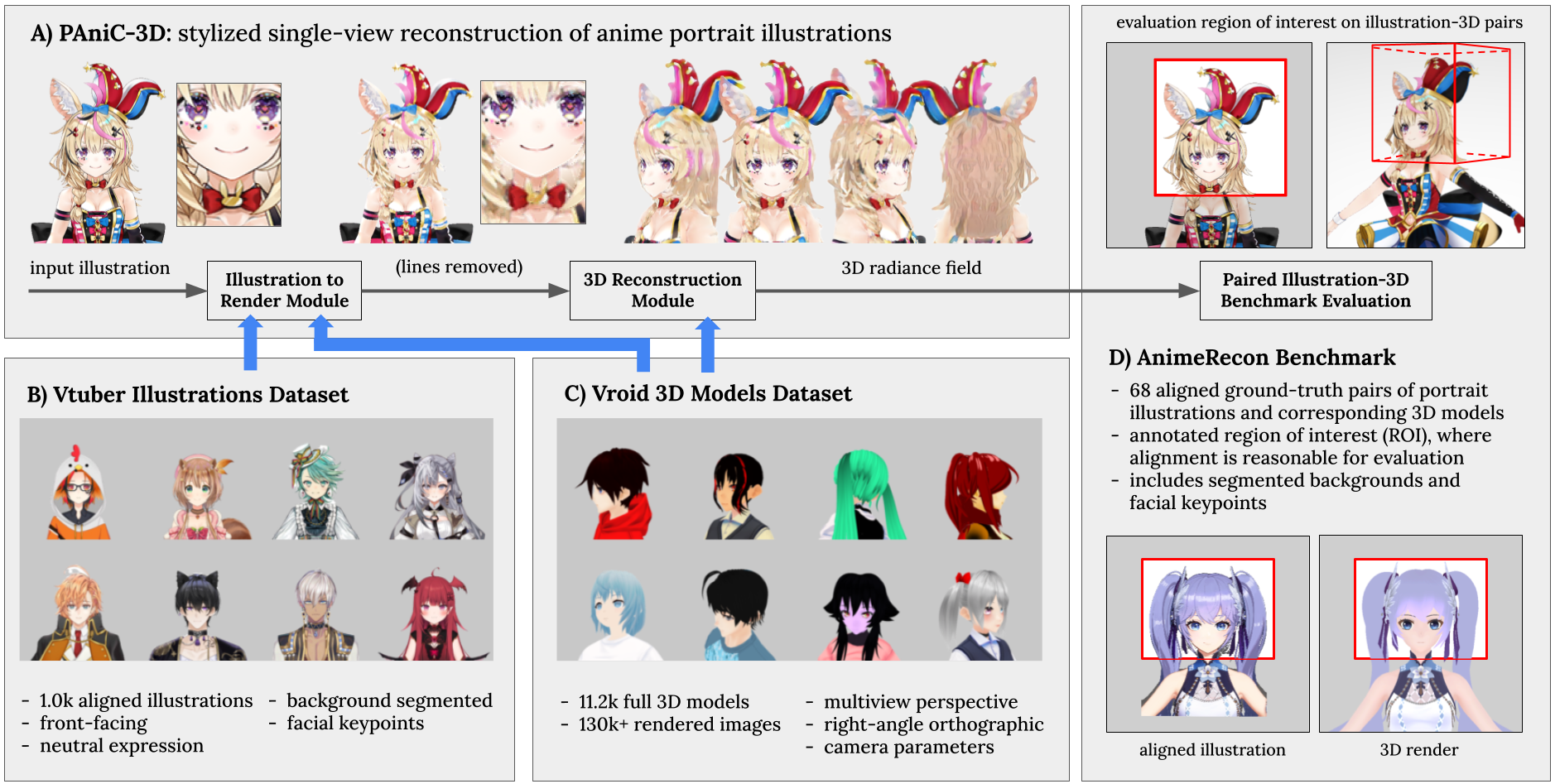}
    \caption{Overview of contributions.  Our (A) PAniC-3D system is able to reconstruct a 3D radiance field directly from a line-based portrait illustration.  We gather a new (B) Vtuber illustration dataset and (C) Vroid 3D models dataset in order to cross the illustration-render domain gap and supervise reconstruction.  To evaluate, we provide a new (D) AnimeRecon benchmark of paired illustrations and 3D models, establishing the novel task of stylized single-view reconstruction of anime characters.  (Art attributions in suppl.)}
    \label{fig:schematic}
\end{figure*}

\section{Introduction \& Related Work}
\label{sec:intro}

With the rise of AR/VR applications, there is increased demand for not only high-fidelity human avatars, but also non-photorealistic 3D characters, especially in the ``anime" style.  Most character designers typically create concept illustrations first, allowing them to express complex and highly diverse characteristics like hair, accessories, eyes, skins, headshapes, etc.  Unfortunately, the process of developing illustrated concept art into an AR/VR-ready 3D asset is expensive, requiring professional 3D artists trained to use expert modeling software.  While template-based creators democratize 3D avatars to an extent, they are often restricted to 3D assets compatible with a specific body model.

We propose PAniC-3D, a system to automatically reconstruct a stylized 3D character head directly from illustrated (p)ortraits of (ani)me (c)haracters.  We formulate our problem in two parts: 1) implicit single-view head reconstruction, 2) from across an illustration-3D domain gap.  To summarize our contributions:
\begin{itemize}
    \item \textbf{PAniC-3D}: a system to reconstruct the 3D radiance field of a stylized character head from a single line-based portrait illustration.
    \item The \textbf{Vroid 3D dataset} of 11.2k character models and renders, the first such dataset in the anime-style domain to provide 3D assets with multiview renders.
    \item The \textbf{Vtuber dataset} of 1.0k reconstruction-friendly portraits (aligned, front-facing, neutral-expression) that bridges the illustration-render domain gap through the novel task of line removal from drawings.
    \item The \textbf{AnimeRecon benchmark} with 68 pairs of aligned 3D models and corresponding illustrations, enabling quantitative evaluation of both image and geometry metrics for stylized reconstruction.
\end{itemize}

\subsection{Implicit 3D Reconstruction}

While there has been much work on mesh-based reconstruction from images \cite{liu2019soft}, these systems are not expressive enough to capture the extreme complexity and diversity of topology of our 3D characters.  Inspired by the recent successes in generating high-quality 3D radiance fields \cite{chan2021efficient,chan2021pi,or2021stylesdf,xue2022giraffe}, we instead turn to implicit representations.  However, to achieve high-quality results, recent implicit reconstruction work such as PixelNerf \cite{pixelnerf} tend to operate solely from 2D images, due to the lack of publicly-available high-quality 3D data.  Some implicit reconstruction systems like Pifu \cite{saito2019pifu} employing complex 3D assets have shown reasonable success using point-based supervision, but require careful point sampling techniques and loss balancing.

There is also a body of work on sketch-based modeling, where 3D representations are recovered from contour images.  For example, Rui \etal \cite{3dsketch} use a multi-view decoder to predict sketch-to-depth and normals, which are then used for surface reconstruction.  Song \etal \cite{profsketch} additionally try to compensate multi-view drawing discrepancies by learning to realign the inputs.   While related to our single-view portrait reconstruction problem, these methods require multi-view sketches that are difficult for character artists to draw consistently, and cannot handle color input.

For our case with complex high-quality 3D assets, we demonstrate the superiority of differentiable volumetric rendering for reconstruction.  We build off of recent unconditional generative work (EG3D \cite{chan2021efficient}), formulating the problem of reconstruction as a conditional generation, proposing several architecture improvements, and applying direct 2.5D supervision signal as afforded by our 3D dataset.

\subsection{Anime-style 3D Avatars and Illustrations}

It is a fairly common task for 3D character artists to produce a 3D model from a portrait illustration; however from the computer graphics standpoint, this stylized reconstruction setup adds additional ambiguity to an already ill-posed problem.  In addition, while there's work in the popular anime/manga domain using 3D character assets (for pose estimation \cite{manpu}, re-targetting \cite{animeceleb,talkinghead}, and reposing \cite{conr}, etc.), there's a lack of publicly-available 3D character assets with multi-view renders that allow scalable training (\cref{tab:datasets3d}).  In light of these issues, we propose AnimeRecon (\cref{fig:schematic}d) to formalize the stylization task with a paired illustration-to-3D benchmark, and provide the Vroid dataset (\cref{fig:schematic}c) of 3D assets to enable large-scale training.

Within the problem of stylized reconstruction, we solve the task of contour removal from illustrations.  There is much work on line extraction \cite{winnemoller2012xdog,mangastruct}, sketch simplification \cite{simo2018mastering,simo2018realink}, reconstruction from lines \cite{dvorovzvnak2020monster,3dsketch}, line exploits for artistic imagery \cite{eisai,zhang2020danbooregion}, and scratch-line removal \cite{scratch0,scratch1,scratch2,scratch3,scratch4}; however, the removal of lines from line-based illustration has seen little focus.  We examine this contour deletion task in the context of adapting drawings to render-like images more conducive to 3D reconstruction; we find that naive image-to-image translation \cite{cyclegan,ugatit} is unsuited to the task, and propose a simple yet effective adversarial training setup with facial feature awareness.  Lastly, we provide a Vtuber dataset (\cref{fig:schematic}b) of portraits to train and evaluate contour removal for 3D reconstruction.

\setlength\intextsep{4pt}
\setlength\tabcolsep{5pt}
\begin{table}[t]
\begin{center}
\begin{tabular}{|l|cc|ccc|}
    \hline
    3D datasets	& Vroid	&	AniRec.	&	AC &	CoNR	&	ADD	\\
    \hline
    3D avail.   	& Y	&	Y	&	-	&	-	&	-	\\
    renders avail.	& Y	&	Y	&	Y	&	-	&	-	\\
    multiview	    & Y	&	Y	&	-	&	Y	&	Y	\\
    paired illustr.	& -	&	Y	&	-	&	-	&	-	\\
    \hline
\end{tabular}
\end{center}
\caption{
    3D anime datasets comparison.  Our new Vroid dataset is the first to make 3D anime models and multiview renders available.  The AniRecon benchmark allows quantitative evaluation of both 2D image and 3D geometry metrics.
    Others left-to-right:
    AnimeCeleb \cite{animeceleb},
    CoNR \cite{conr},
    Anime Drawings Dataset \cite{manpu}.
}
\label{tab:datasets3d}
\end{table}

\setlength\intextsep{4pt}
\setlength\tabcolsep{8pt}
\begin{table}[t]
\begin{center}
\begin{tabular}{|l|c|cccc|}
    \hline
    Portraits	& Vtuber	&	AC	&	AP &	iCF	&	BP	\\
    \hline
    illustrations	& Y	&	-	&	Y	&	Y	&	Y	\\
    aligned face	& Y	&	Y	&	Y	&	Y	&	-	\\
    front-facing	& Y	&	Y	&	-	&	-	&	-	\\
    neutral expr.	& Y	&	Y	&	-	&	-	&	-	\\
    segmented   	& Y	&	Y	&	-	&	-	&	Y	\\
    face kpts.  	& Y	&	-	&	-	&	-	&	Y	\\
    \hline
\end{tabular}
\end{center}
\caption{
    Anime image datasets comparison.  Our new Vtuber dataset allows us to examine the domain gap between line-based illustrations and 3D renders, and is filtered/standardized specifically for characteristics desirable in 3D head reconstruction. 
    Others left-to-right:
    AnimeCeleb \cite{animeceleb},
    AnimePortraits \cite{gwernportraits},
    iCartoonFace \cite{icartoonface},
    BizarrePose \cite{bizarre}.
}
\label{tab:datasets2d}
\end{table}



\section{Methodology}
\label{sec:method}


PAniC-3D is composed of two major components (\cref{fig:schematic}a): a 3D reconstructor directly supervised to predict a radiance field from a given front render, and an illustration-to-render module that translates images to the reconstructor's training distribution.  The two parts are trained independently, but are used sequentially at inference.

\subsection{3D Reconstruction Module}
\label{sec:reconmodule}

\begin{figure*}[t]
    \centering
    \includegraphics[width=1\linewidth]{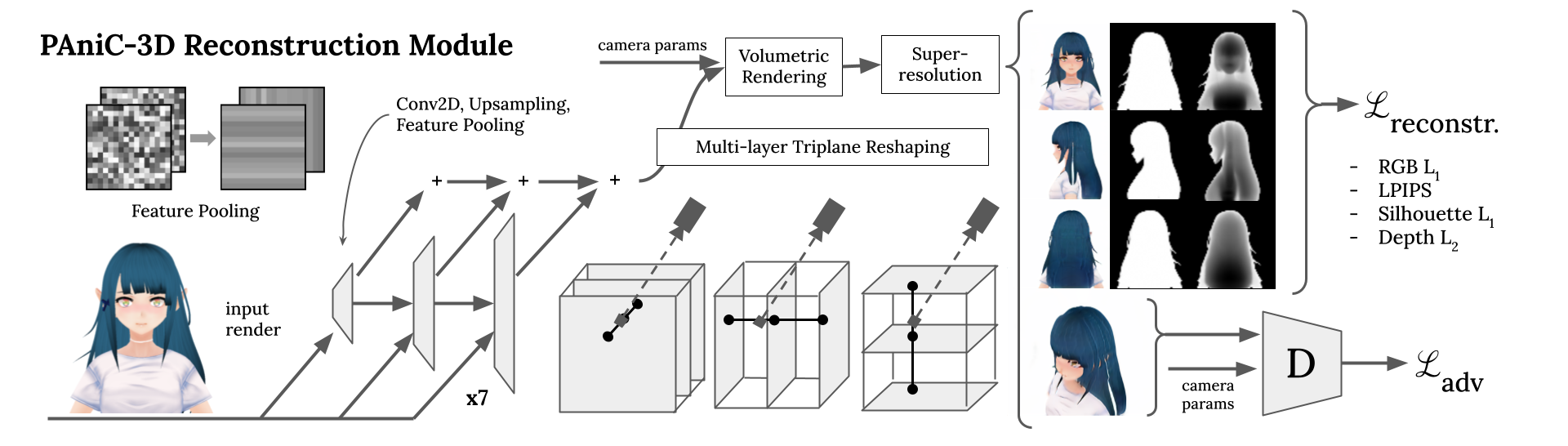}
    \caption{Schematic of the 3D reconstruction module.  The front-orthographic input rendering is fed to a series of upsampling convolutions, with intermediate feature pooling to help distribute information along common triplane axes.  The final feature stack is reshaped into a multi-layer triplane, which is volumetrically rendered and super-resolutioned to the final output.  Reconstruction losses are applied to front, left, right, and back views (right view omitted in figure), and adversarial loss is applied to a random perspective view.  (Art attributions in suppl.)}
    \label{fig:reconschematic}
\end{figure*}

The 3D reconstruction module \cref{fig:reconschematic} is trained with direct supervision to invert a frontal render into a volumetric radiance field.  We build off of recent unconditional generative work (EG3D \cite{chan2021efficient}), formulating the problem of reconstruction as that of conditional generation, proposing several architecture improvements, and applying direct 2.5D supervision signal as afforded by our 3D dataset.

\textbf{Conditional input:}  The given front orthographic view to reconstruct is resized and appended to the intermediate feature maps of the Stylegan2 backbone used in EG3D \cite{chan2021efficient}.  In addition, at the earliest feature map we give the model high-level domain-specific semantic information about the input by concatenating the penultimate features of a pre-trained Resnet-50 anime tagger.  The tagger provides high-level semantic features appropriate for conditioning the generator; it was pretrained by prior work \cite{bizarre} on 1062 relevant classes such as blue\_hair, cat\_ears, twin\_braids, etc.

\textbf{Feature pooling:}  As the spatial feature maps are to be reshaped into a 3D triplane as in EG3D \cite{chan2021efficient}, we found it beneficial to pool a fraction of each feature map's channels along the image axes (see \cref{fig:reconschematic} left).  This simple technique helps distribute information along common triplane axes, improving performance on geometry metrics.

\textbf{Multi-layer triplane:}  As proposed in concurrent work \cite{panohead}, we improve the EG3D triplane by stacking more channels along each plane (see \cref{fig:reconschematic} center).  The method may be interpreted as a hybrid between a triplane and a voxel grid (they are equivalent if the number of layers equals the spatial size).  Setting three layers per plane allows better spatial disambiguation when bilinearly sampling the volume, and particularly helps our model generate more plausible backs of heads (a challenge not faced by EG3D).

\textbf{Losses:}  We take full advantage of the ground-truth 2.5D representations afforded to us by our available 3D assets.  Our reconstruction losses include: RGB $L_1$, LPIPS \cite{lpips}, silhouette $L_1$, and depth $L_2$; these are applied to the front, back, right, and left orthographic views, as shown in \cref{fig:reconschematic}.  A discriminative loss is applied to improve the detail quality, in addition to maintaining the generation orientation.  We also keep the R1 and density regularization losses from EG3D training.  Our 2.5D representations and adversarial setup allow us to surpass similar single-view reconstructors such as PixelNerf \cite{pixelnerf} which work only with color losses.

\textbf{Post-processing:}  We leverage our assumption that front-orthographic views are given as input, by stitching the given input onto the generated radiance field at inference.  The xy-coordinates of each pixel's intersection within the volume are used to sample the input as a uv-texture map; we cast few additional rays from each intersection to test for visibility from the front, and apply the retexturing accordingly.  This simple yet effective method improves detail preservation from the input at negligible cost.

\subsection{Illustration-to-Render Module}
\label{sec:img2module}

\begin{figure}[t]
    \centering
    \includegraphics[width=1\linewidth]{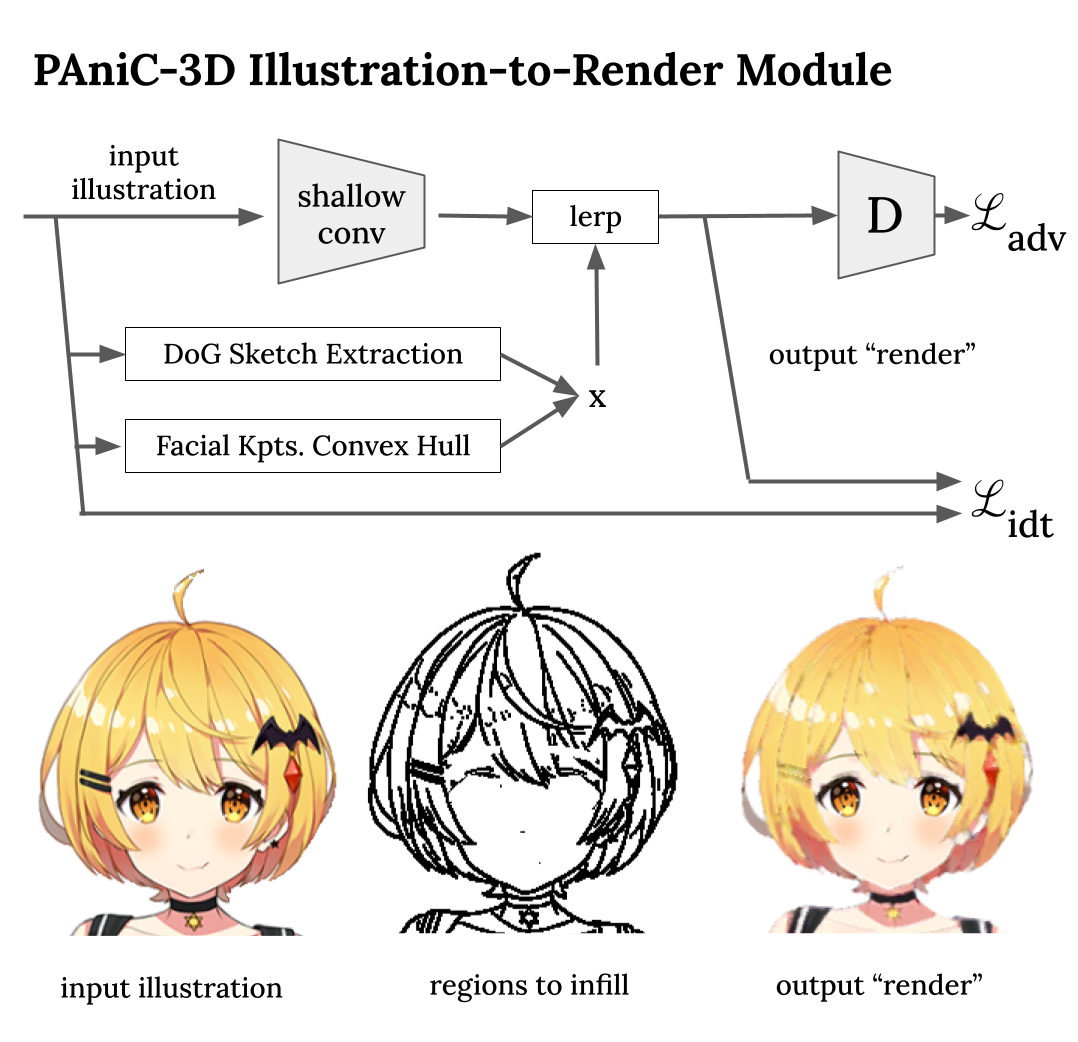}
    \caption{Schematic of the illustration-to-render module.  We design a simple yet effective network to cross the domain gap by removing illustration contour lines absent in a diffuse 3D render, while retaining lines present around facial features like the eyes and mouth. (Art attributions in suppl.)}
    \label{fig:img2imgschematic}
\end{figure}

In order to remove non-realistic contour lines present in the input illustration, but absent in a diffusely-lit radiance field, we design an illustration-to-render module (\cref{fig:img2imgschematic}).  Assuming access to unpaired illustrations and renders (our Vtuber and Vroid datasets, respectively), the shallow network re-generates pixel colors near lines in the drawing in order to adversarially match the render image distribution.

Similar to unpaired image-to-image models like CycleGAN and UGATIT \cite{cyclegan,ugatit}, we also impose a small identity loss; while this may seem counter-productive for our infilling case where identity is preserved in non-generated regions, we found that this stabilizes the GAN training. Note that our setup also differs from other infilling models, in that we inpaint to match a distribution different from the input.

Following prior work extracting sketches from line-based animations \cite{eisai}, we use the simple difference of gaussians (DoG) operator in order to prevent double-line extraction around each stroke.

While most lines present in the drawing should be removed, certain lines around key facial features must be preserved as they indeed appear in renderings (eyes, mouth, nose, etc.).  We employ an off-the-shelf anime facial landmark detector \cite{hysts} to create convex hulls around critical structures, where infilling is disallowed.

\begin{figure}[t]
    \centering
    \includegraphics[width=1\linewidth]{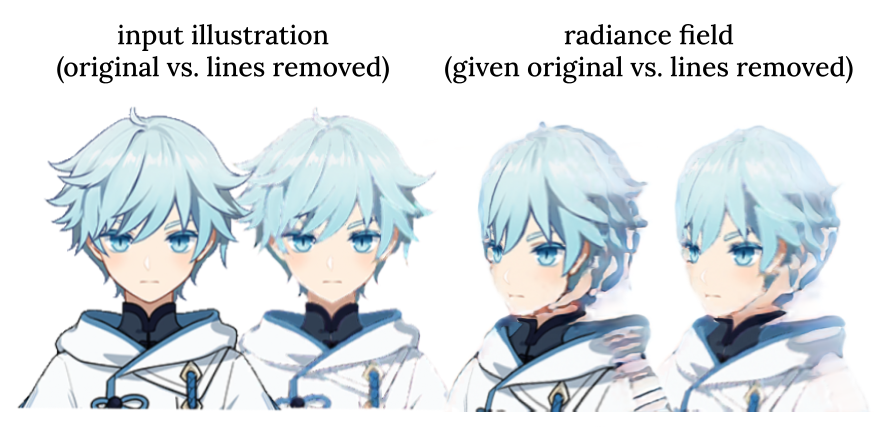}
    \caption{The effect of our illustration-to-render module on reconstruction.  Without our proposed line-infilling method, the reconstruction module trained on contour-less renders is unable to cross the illustration domain gap.  Notice the line artifacts along the shoulders, and improper contours along the chins.  (Art attributions in suppl.)}
    \label{fig:img2imgradfield}
\end{figure}

We show that this line removal module indeed achieves a more render-like look; it performs image translation more accurately than baseline methods when evaluated over our AnimeRecon pairs (\cref{tab:img2imgresults}), and removes line artifacts from the ultimate radiance field renders (\cref{fig:img2imgradfield}).

\begin{figure}[tp]
    \centering
    \includegraphics[width=1\linewidth]{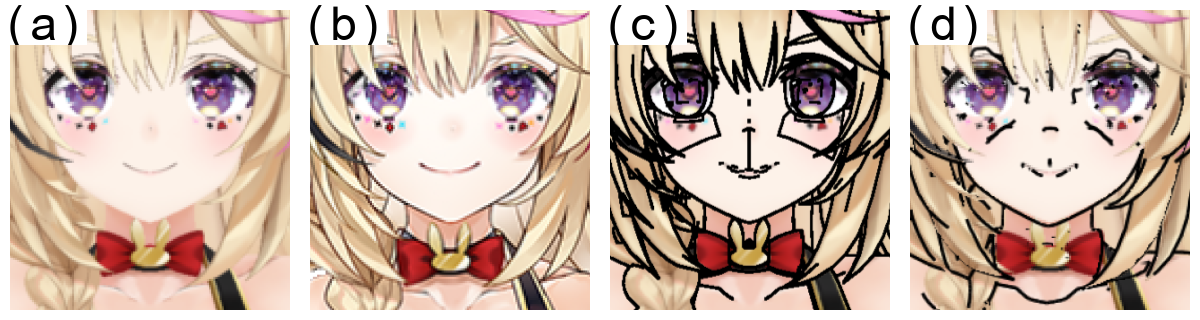}
    \caption{(a) No-line diffuse render, (b) real-lined illustration, (c) Blender Freestyle \cite{freestyle}, (d) RTSC suggestive contours \cite{suggcontour}.  Toon shaders over-draw (c, cheeks) or miss lines (d, bowtie); it is non-trivial to model the artistic line placement in real drawings (b).  Thus, we train our Illustration-to-Render module on real lines drawn by artists instead of synthetically-added lines.}
    \label{fig:suggcontour}
\end{figure}

\begin{figure*}[t]
    \centering
    \includegraphics[width=1\linewidth]{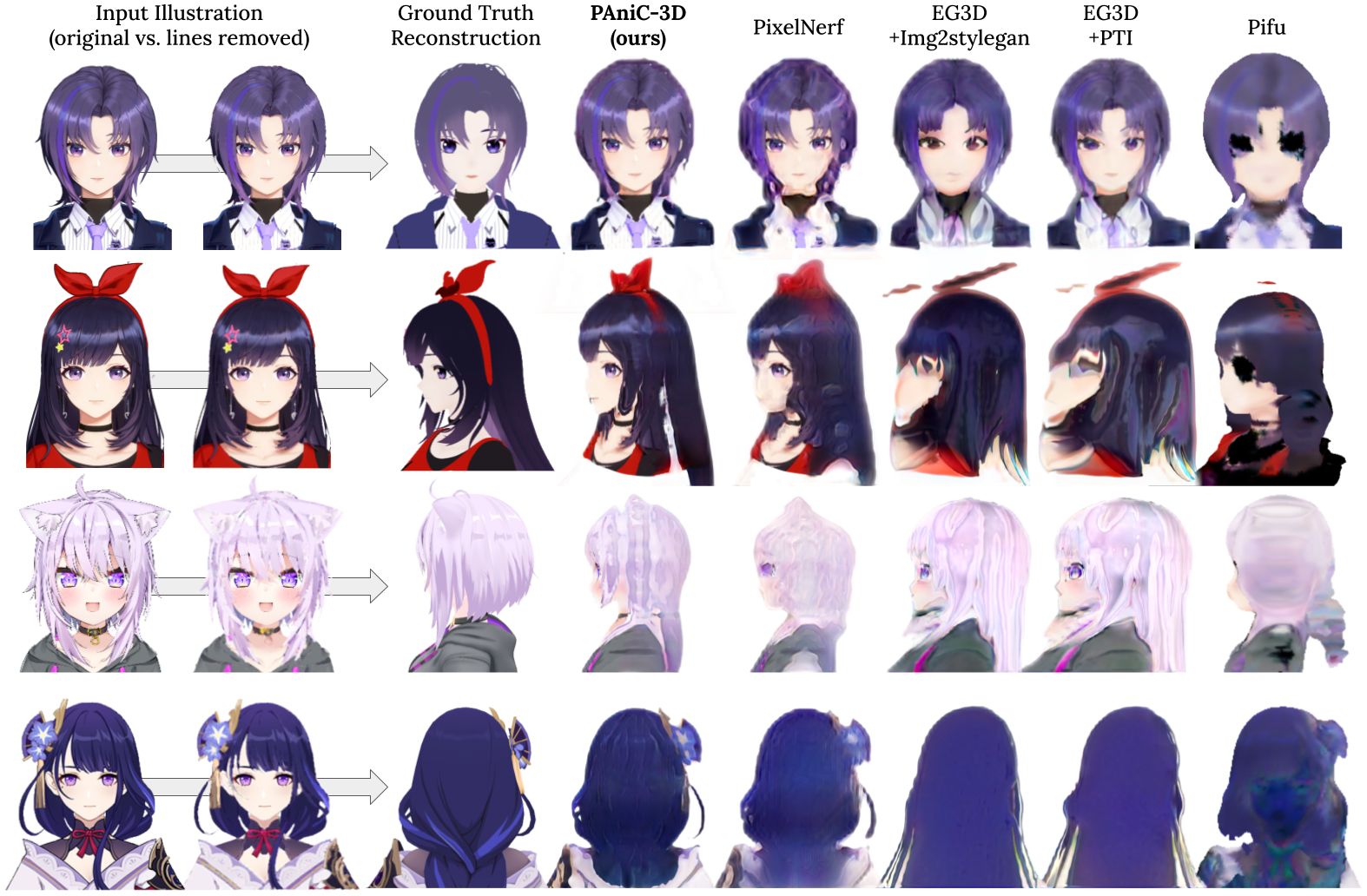}
    \caption{Qualitative comparison of baselines.  Illustrations with lines removed by our illustration-to-render module are fed to various reconstruction frameworks; our PAniC-3D system delivers plausible reconstructions, while other methods struggle to preserve identity and predict reasonable geometry.  Note that metrics (\cref{tab:baselines}) between the displayed ground-truth and prediction are restricted to an ROI bounding box/rectangular prism (unshown) during evaluation.  (Art attributions in suppl.)}
    \label{fig:baselines}
\end{figure*}


\section{Data}
\label{sec:data}


Unless otherwise mentioned, we use 80-10-10 splits for training, validation, and testing.

\subsection{Vroid 3D Dataset}
\label{sec:vroid}

We collect a large dataset of 11.2k 3D anime characters from VroidHub, in order to train both the reconstruction and image-to-image translation modules of PAniC-3D.  Unlike previous work in the 3D anime domain using MikuMikuDance PMD/PMX models \cite{manpu,animeceleb,talkinghead}, Vroid VRM models conform to the GLTF2 standard \cite{gltf2} with several extensions \cite{vrm}, allowing us to render using ModernGL \cite{moderngl,mglw}. As the data is crowd-sourced, we filter against a variety of undesirable properties, such as: texture corruption, too much or too little transparency, extreme character sizes, missing bones, etc.  The 11.2k renders we use represent 70\% retension of our original 16.0k scraped models.

All our image data are rendered with unit ambient lighting (\eg using diffuse surface color only), unit distance away from the ``neck" bone common to VRMs \cite{vrm}.  Our choice to model only diffuse lighting is motivated by the frequent artistic decision to paint speculars as textures (\cref{tab:img2imgresults} row 2, hair and eye highlights on diffuse renders).  Adding specular renderering to painted highlights would introduce inconsistent lighting effects.  Linear blend skinning is used to lower characters' arms 60-degrees from their resting T-pose position.  We supersample at 1024px resolution, before bilinear downsampling to 512px for training and testing.

The dataset for PAniC-3D's reconstruction module consists of both random perspective views for adversarial training, and fixed orthographic views for the input and reconstruction losses.  Each 3D model is rendered from 8 random perspective views (with uniformly-sampled 360-degree azimuth, normally-sampled elevation of 20-degree standard deviation, and fixed 30-degree full field-of-view); this yields a total of 89.6k perspective images, each with known camera parameters.  The four orthographic views are taken at fixed 90-degree angles from the front, sides, and back.  The unpaired 3D data for our image-to-image translation module is simply the front orthographic view of each character identity.

\subsection{Vtuber Portraits Dataset}
\label{sec:vtuber}

We gather 1004 portraits from the VirtualYoutuber Fandom Wiki as the unpaired illustration dataset for our image-to-image module.  These portraits were manually filtered from 15.2k scraped images for desired properties, \eg high-resolution, front-facing, neutral-expression, uncropped, etc.  In order to mimic the image distribution of the orthographic front-view Vroid renders, we white out the backgrounds using the character segmenter from Chen \etal \cite{bizarre} and select the largest connected component.  In addition, we align the facial keypoints of each illustration (extracted with a pretrained YOLOv3 \cite{hysts,yolo3}) to match the height and scale distributions of keypoint detections on the Vroid renders.

While we could add lines artificially, we found toon shaders \cite{freestyle,suggcontour} to poorly model artistic judgements needed to place lines on intricate characters (\cref{fig:suggcontour}).  To avoid these artifacts, we designed a data-driven Illustr2Render module to train on real lines drawn by artists.

\subsection{AnimeRecon Benchmark}
\label{sec:animerecon}

We collect a benchmark set for stylized reconstruction by finding 68 characters with both 3D models and closely-aligned illustrations.  Specifically, we source from the 3D mobile game Genshin Impact (for which we can match character portraits from the Fandom Wiki) and the virtual talent agency Hololive (from which several members have both 3D avatars and a Fandom Wiki portrait).  As the raw 3D data from both sources comes in MMD format, we convert to VRM using the DanSingSing converter; the rendering process is the same as that of Vroid.

The portraits are aligned to their corresponding front-view orthographic renders by a manually-decided mixture of YOLOv3 facial keypoints \cite{hysts,yolo3} and ORB detections.  We segment out backgrounds using the same model \cite{bizarre} and procedure as with Vtuber portraits.  In order to maintain separation from training data, we remove all Hololive identities present in this benchmark set from the Vtuber portraits set.

Inevitably, the illustrations do not all align perfectly with their corresponding renders, and many Genshin illustrations are cropped when aligned.  In order to perform more reasonable evaluation, we additionally label a rectangular region of interest (ROI) for each aligned pair, within which the alignment is appropriate; all 2D image and 3D geometry metrics reported are restricted to the ROI when calculated.

\setlength\intextsep{4pt}
\setlength\tabcolsep{4pt}
\begin{table*}[t]
\begin{center}
\begin{tabular}{|l||rrr|rrr|rrr|rrr|}
	\hline																								
	&		&	front	&		&		&	back	&		&		&	360	&		&		&	geom.	&		\\
	&	CLIP	&	LPIPS	&	PSNR	&	CLIP	&	LPIPS	&	PSNR	&	CLIP	&	LPIPS	&	PSNR	&	CD	&	F-1@5	&	F-1@10	\\
	\hline																								
PAniC-3D (full)	&	\textbf{94.66}	&	\textbf{19.37}	&	16.91	&	\textbf{85.08}	&	\textbf{30.02}	&	15.51	&	\textbf{84.68}	&	{25.25}	&	\textbf{15.98}	&	\textbf{1.33}	&	\textbf{37.73}	&	65.50	\\
no feature pooling	&	\textbf{94.64}	&	\textbf{19.26}	&	\textbf{16.95}	&	84.06	&	\textbf{30.23}	&	\textbf{15.53}	&	84.30	&	\textbf{25.42}	&	15.96	&	1.38	&	35.26	&	65.51	\\
no multi-layer triplane	&	94.39	&	19.99	&	{16.94}	&	84.28	&	30.37	&	15.41	&	{84.49}	&	26.13	&	15.82	&	1.44	&	34.55	&	\textbf{65.95}	\\
no side/back loss	&	94.54	&	19.93	&	16.86	&	\textbf{85.50}	&	32.18	&	14.73	&	83.98	&	27.34	&	15.39	&	1.56	&	36.82	&	58.67	\\
no 2.5D loss	&	{94.10}	&	{20.76}	&	\textbf{17.72}	&	{83.27}	&	32.51	&	15.11	&	\textbf{84.58}	&	26.19	&	15.65	&	{1.37}	&	36.42	&	63.88	\\	
	\hline																								
PixelNerf \cite{pixelnerf}	&	91.07	&	22.96	&	16.76	&	81.02	&	32.01	&	\textbf{15.74}	&	80.42	&	\textbf{24.86}	&	\textbf{16.31}	&	1.45	&	35.07	&	65.50	\\
																									
EG3D+Img2SG \cite{abdal2019image2stylegan}	&	85.90	&	30.78	&	13.92	&	77.78	&	39.84	&	12.68	&	79.78	&	31.10	&	13.86	&	2.05	&	11.88	&	23.55	\\
EG3D+PTI \cite{pti}	&	89.90	&	25.93	&	15.50	&	77.37	&	39.16	&	13.37	&	79.78	&	32.62	&	14.32	&	2.19	&	11.38	&	23.54	\\
																									
Pifu \cite{saito2019pifu}	&	75.12	&	41.62	&	11.94	&	75.01	&	43.63	&	12.51	&	73.62	&	32.45	&	13.81	&	\textbf{1.35}	&	\textbf{37.37}	&	\textbf{68.13}	\\
	\hline																								
\end{tabular}
\end{center}
\caption{
    Ablations and baselines.  We evaluate and compare our system using our new AnimeRecon benchmark of illustrations with their ground-truth 3D reconstructions.  Image reconstruction metrics are listed for the front (reconstruction of the given input) and back orthographic views, as well as averaged over twelve perspective spin-around views.  On the right, chamfer distances and F-1 scores \cite{meshrcnn} capture the correctness of reconstructed meshes.  For fair comparison, lines are preprocessed out of all input images using our illustration-to-render model, and evaluation is performed within our annotated ROIs.  The top two of each category are bolded.
}
\label{tab:baselines}
\end{table*}


\begin{figure}[t]
    \centering
    \includegraphics[width=1\linewidth]{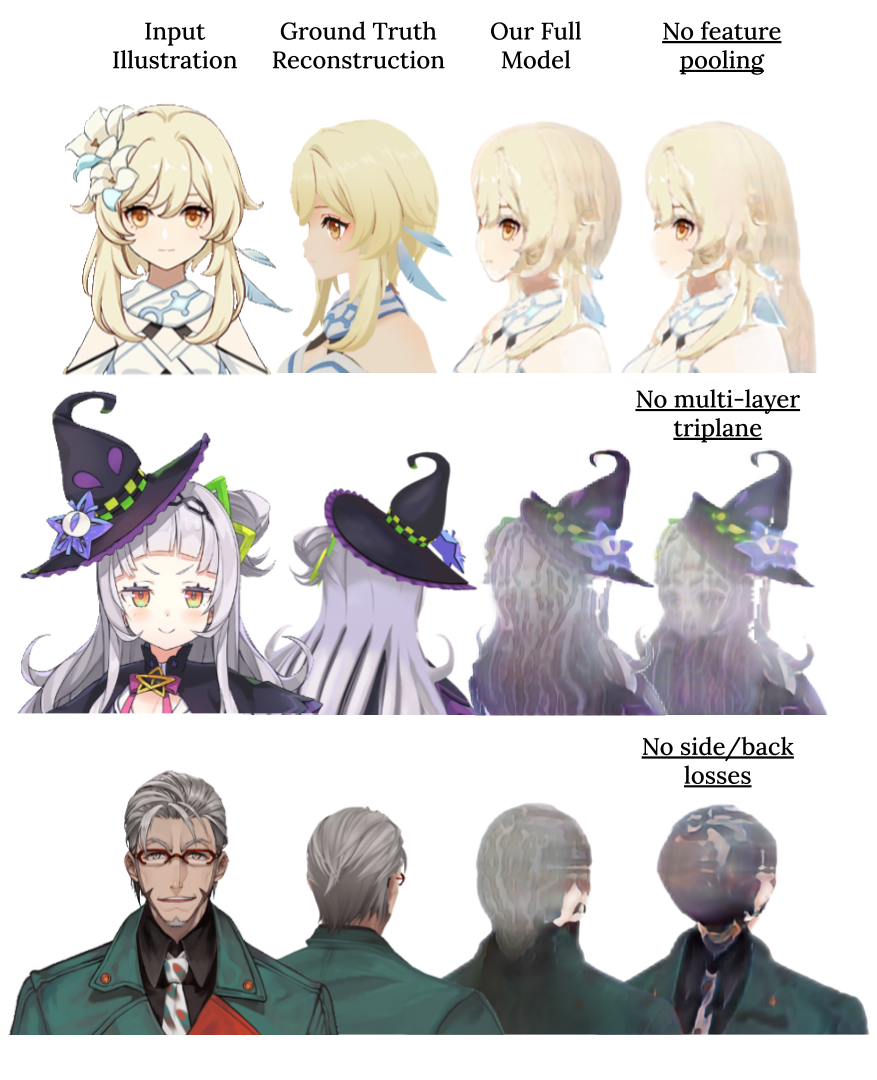}
    \caption{3D reconstruction module ablations. Row 1: feature pooling helps distribute information across the triplane, and correctly structures the hair as short. Rows 2+3: the multi-layer triplane helps disambiguate the front/back, and side/back-view losses significantly improve both geometry and texture. (Art attributions in suppl.)}
    \label{fig:ablations}
\end{figure}

\section{Results \& Evaluation}
\label{sec:results}

In this section, we provide a breakdown of reconstruction and image-to-image translation performance, with both qualitative and quantitative comparison to other baselines.

\subsection{3D Reconstruction Results}
\label{sec:reconresults}

\subsubsection{Metrics}

As shown in \cref{tab:baselines}, we evaluate over our new AnimeRecon benchmark using both 2D image and 3D geometry metrics.  All the illustration inputs went through our illustration-to-render module before being fed to the respective method; we believe this is a fairer comparison, as all the reconstructors were trained on renders.

Image metrics are measured by comparing the predicted radiance field's integrated image with the ground-truth render from the same camera viewpoint.  We show such measures for the front orthographic view (which should match the input), the back orthographic view, and an average of 12 perspective cameras circling the character at 30-degree intervals.  For the front and back views, we restrict evaluation to our AnimeRecon ROI (\cref{fig:schematic}d); for the 12 circling views, we crop to the horizontal bounding box strip.

We show standard color metrics like PSNR for completeness, but as there are inevitably imperfections on the AnimeRecon illustration-render pairs (even within the ROI), perceptual metrics like CLIP image cosine similarity \cite{clip} and LPIPS \cite{lpips} are generally more relevant to quality.

The geometry metrics are on the right of \cref{tab:baselines}.  We extract meshes from both the ground-truth 3D asset and the predicted radiance field (through marching cubes), and delete faces with vertices outside the rectangular prism defined by the ROI annotation (see \cref{fig:schematic}d top).  We then sample 10k points randomly from each mesh subset, to compute the point-cloud chamfer distance and F-1 scores at 5cm and 10cm \cite{meshrcnn}.  To put F-1 in perspective, the average Vroid head width is 25.5cm (real heads are 14cm).  The F-1 are low overall, due to larger proportions and protruding hair/accessories with large surface area.

Though Mesh R-CNN also proposes the normal consistency metric \cite{meshrcnn}, it is difficult to obtain for our data, as there are assets with broken ground truth normals (zero values, inverted meshes).  We can calculate rough consistency by re-estimating all normals, resulting in: Ours 75.6, PixelNerf 77.5, Img2SG 73.0, PTI 72.7, Pifu 77.3.  We observed that these approximate measures don't match well with visual quality, but include them here for completeness.

\subsubsection{Baselines}

Comparisons are shown between our method and several other implicit reconstruction methods (\cref{fig:baselines}).  Of the methods shown, only Pifu \cite{saito2019pifu} receives point-wise signals for optimization; we see that this works to its detriment, as the near-surface point sampling strategies bias the model towards certain geometric structures (such as black eyelashes that often protrude from the face).  The other methods using volumetric rendering inherently weigh the supervision signal such that the final rendered product is consistent.

Image2stylegan \cite{abdal2019image2stylegan} and Pivotal Tuning Inversion \cite{pti} are optimization-based methods of performing single-view reconstruction, the latter of which was used in EG3D to demonstrate reconstruction by projection \cite{chan2021efficient}.  We train an EG3D with similar hyperparameters as the Stylegan2 backbone used in our reconstruction module, and allow the two respective baselines to optimize features/weights in the prior to match the given input.  Unfortunately, the EG3D prior is not sufficient to regularize the optimization, leading to implausible results; while PTI worked reasonably well for EG3D, it may not work as well on our setup where the back of the head must also be projected.

Lastly we compare to our model with the PixelNerf \cite{pixelnerf} single-view setup.  The key difference in this comparison is that PixelNerf does not use adversarial nor 2.5D reconstruction losses (for fairer comparison with us, we trained PixelNerf with LPIPS in addition to $L_1$, resulting in significantly less blurry results).  We see that without these signals provided by our available 3D assets, the quality of details and geometry does not match up to PAniC-3D.

\subsubsection{Ablations}

From \cref{tab:baselines} and \cref{fig:ablations}, we conclude that our architecture decisions improve both the qualitative and quantitative performance of our reconstruction system.  It is shown that feature pooling is able to propagate information along triplane axes to improve generated geometries, the multi-layer triplane adds additional model capacity to further disambiguate locations for features, and the addition of fixed side/back-view losses significantly improves geometry and texture.  Expectedly, removing 2.5D supervision also deteriorates performance.

\subsection{Illustration-to-Render Results}
\label{sec:img2results}

As shown in \cref{tab:img2imgresults} and \cref{fig:img2imgradfield}, our illustration-to-render model was able to effectively remove contours that would not appear in renders.  We evaluate in \cref{tab:img2imgresults} how closely the translated illustrations match their rendered counterpart from the AnimeRecon benchmark (restricted to the ROI annotation), and find that we are able to significantly outperform naive inpainting \cite{telea} as well as off-the-shelf image-to-image translation systems \cite{ugatit,cyclegan}.  The others struggled to retain key semantic structures like eyes/mouths, and often failed to retain the original identity.

In \cref{fig:img2imgradfield}, it is shown that removing lines indeed alleviates the effects of domain transfer between illustrations and their output predicted radiance field; artifacts like extra contours and bands along shoulders are significantly reduced using our line removal strategy.

\setlength\intextsep{4pt}
\setlength\tabcolsep{8pt}
\begin{table}[t]
\begin{center}
    \includegraphics[width=1\linewidth]{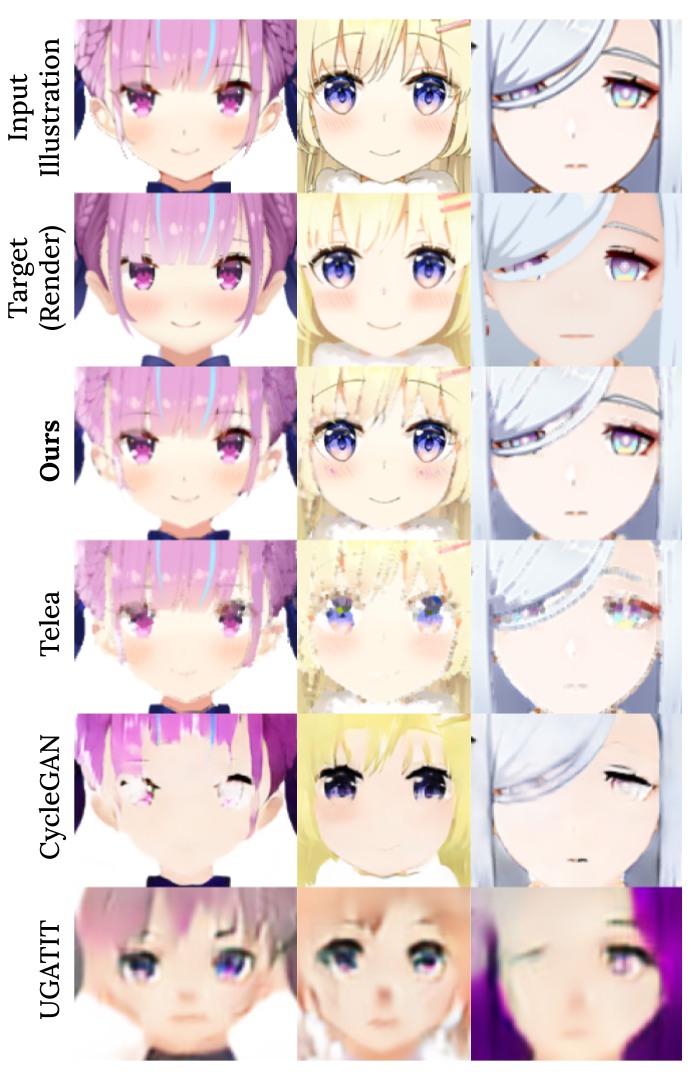}
\begin{tabular}{|l|ccc|}
    \hline
     	& LPIPS↓	&	CLIP↑	&	PSNR↑ 	\\
    \hline
    Ours   	                    & \textbf{18.26}	&	\textbf{94.97}	&	\textbf{16.96}	\\
    Telea \cite{telea}	        & 23.91	&	93.88	&	16.67	\\
    CycleGAN \cite{cyclegan}    & 21.39	&	93.81	&	15.23	\\
    UGATIT \cite{ugatit}    	& 39.64	&	85.48	&	13.18	\\
    \hline
\end{tabular}
\end{center}
\caption{
    Illustration-to-render results.  Over our AnimeRecon benchmark, we calculate perceptual and color metrics between translated illustrations and their corresponding 3D renders, restricted to the ROI annotations.  Our method crosses the illustration-render domain gap better than both naive inpainting (which struggles to retain facial features) and deep image2image models (which fail to retain identity).  LPIPS is scaled by 1e2.
}
\label{tab:img2imgresults}
\end{table}



\section{Limitations \& Future Work}
\label{sec:conclusion}

From the figures comparing the ground-truth 3D model renders and generated radiance fields, it is evident that this task is incredibly difficult.  Although PAniC-3D performs better than other baseline methods, there is still a large gap in quality between our reconstructions and real character assets; this is still particularly evident for the hind occluded regions, as well as the areas around the ears connecting the front and back.  Our model is usually able to prevent faces from appearing on the backside, but still largely relies on copying the visible portions of accessories to cover the occluded parts.  In the future, we may look into ways of decomposing the radiance field in an object-centric manner, and generate accessories through a separate more expressive process.  An obvious extension of this work would be to model full-body characters and exploit more opportunities given by the Vroid dataset, which supports blendshape facial expressions, hair physics, full-body rigging, and more.

In conclusion, we propose PAniC-3D, a system to reconstruct stylized 3D character heads directly from illustrated portraits of anime characters.  Our anime-style domain poses unique challenges to single-view reconstruction when compared to natural images of human heads, such as hair and accessories with more complex and diverse geometry, and non-photorealistic contour lines.  Furthermore, there is a lack of both 3D model and portrait illustration data suitable to train and evaluate this ambiguous stylized reconstruction task.  Facing these challenges, our proposed PAniC-3D architecture crosses the illustration-to-3D domain gap with a line-filling model, and represents sophisticated geometries with a volumetric radiance field.  We train our system with two large new datasets (11.2k Vroid 3D models, 1k Vtuber portrait illustrations), and evaluate on a novel AnimeRecon benchmark of illustration-to-3D pairs.  PAniC-3D significantly outperforms baseline methods, and can generate plausible fully-textured geometries from a single input drawing.  We hope that our proposed system and provided datasets may help establish the stylized reconstruction task.


\section*{Acknowledgements}

The authors would like to thank Softmind Ltd. for sharing their DanSingSing PMX-to-VRM converter, Geng Lin for adding functions to the converter, Srinidhi Hegde for valuable discussion on related topics, and NVIDIA for additional GPU support.

{\small
\bibliographystyle{ieee_fullname}
\bibliography{egbib}
}

\end{document}